\documentclass[10pt,twocolumn,letterpaper]{article}

\usepackage{wacv}
\usepackage{times}
\usepackage{epsfig}
\usepackage{graphicx}
\usepackage{amsmath}
\usepackage{amssymb}
\usepackage{bm}
\usepackage{float}

\newenvironment{packed_enum}{
\begin{itemize}
  \setlength{\itemsep}{3pt}
  \setlength{\parskip}{0pt}
  \setlength{\parsep}{0pt}
}{\end{itemize}}

\usepackage{mathtools}
\DeclarePairedDelimiter\abs{\lvert}{\rvert}%
\makeatletter
\let\oldabs\abs
\def\abs{\@ifstar{\oldabs}{\oldabs*}}
\DeclarePairedDelimiter\norm{\lVert}{\rVert}%
\let\oldnorm\norm
\def\norm{\@ifstar{\oldnorm}{\oldnorm*}}
\makeatother

\usepackage[pagebackref=true,breaklinks=true,letterpaper=true,colorlinks,bookmarks=false]{hyperref}

\wacvfinalcopy 

\def\wacvPaperID{432} 

\begin{document}

\title{Fully Convolutional Cross-Scale-Flows for Image-based Defect Detection}

\author{Marco Rudolph\textsuperscript{1}
\and
Tom Wehrbein\textsuperscript{1}
\and
Bodo Rosenhahn\textsuperscript{1}
\and
Bastian Wandt\textsuperscript{2}
\and
\hspace{-2.5mm}\textsuperscript{1}Leibniz University Hannover, Germany \quad \quad \textsuperscript{2}University of British Columbia, Canada\\
{\tt\small rudolph@tnt.uni-hannover.de}
}

\maketitle

\maketitle

\begin{abstract}
In industrial manufacturing processes, errors frequently occur at unpredictable times and in unknown manifestations.
We tackle the problem of automatic defect detection without requiring any image samples of defective parts.
Recent works model the distribution of defect-free image data, using either strong statistical priors or overly simplified data representations.
In contrast, our approach handles fine-grained representations incorporating the global and local image context while flexibly estimating the density.
To this end, we propose a novel fully convolutional cross-scale normalizing flow (CS-Flow) that jointly processes multiple feature maps of different scales.
Using normalizing flows to assign meaningful likelihoods to input samples allows for efficient defect detection on image-level.
Moreover, due to the preserved spatial arrangement the latent space of the normalizing flow is interpretable
which enables to localize defective regions in the image.
Our work sets a new state-of-the-art in image-level defect detection on the benchmark datasets Magnetic Tile Defects and MVTec AD
showing a 100\% AUROC on 4 out of 15 classes.
\end{abstract}
\vspace{-2mm}
\section{Introduction}

During the industrial production of components, defects occur over time.
They must be detected to ensure safety standards and product quality.
Since manual inspection by humans is very costly and error-prone, reliable and efficient automatic defect detection is highly demanded.
In most real-world scenarios, however, there exist no examples of such defects.
Moreover, even if a small set of known defects is available, new and formerly unseen types of defects occur at unpredictable times which makes it impossible to apply standard classification approaches.
Instead, it is inevitable to let the defect detector learn only from non-defective examples.
This problem is commonly called semi-supervised anomaly detection (AD), novelty detection or one-class classification.
\begin{figure}[h!]
\centering
  \includegraphics[width=0.5\textwidth]{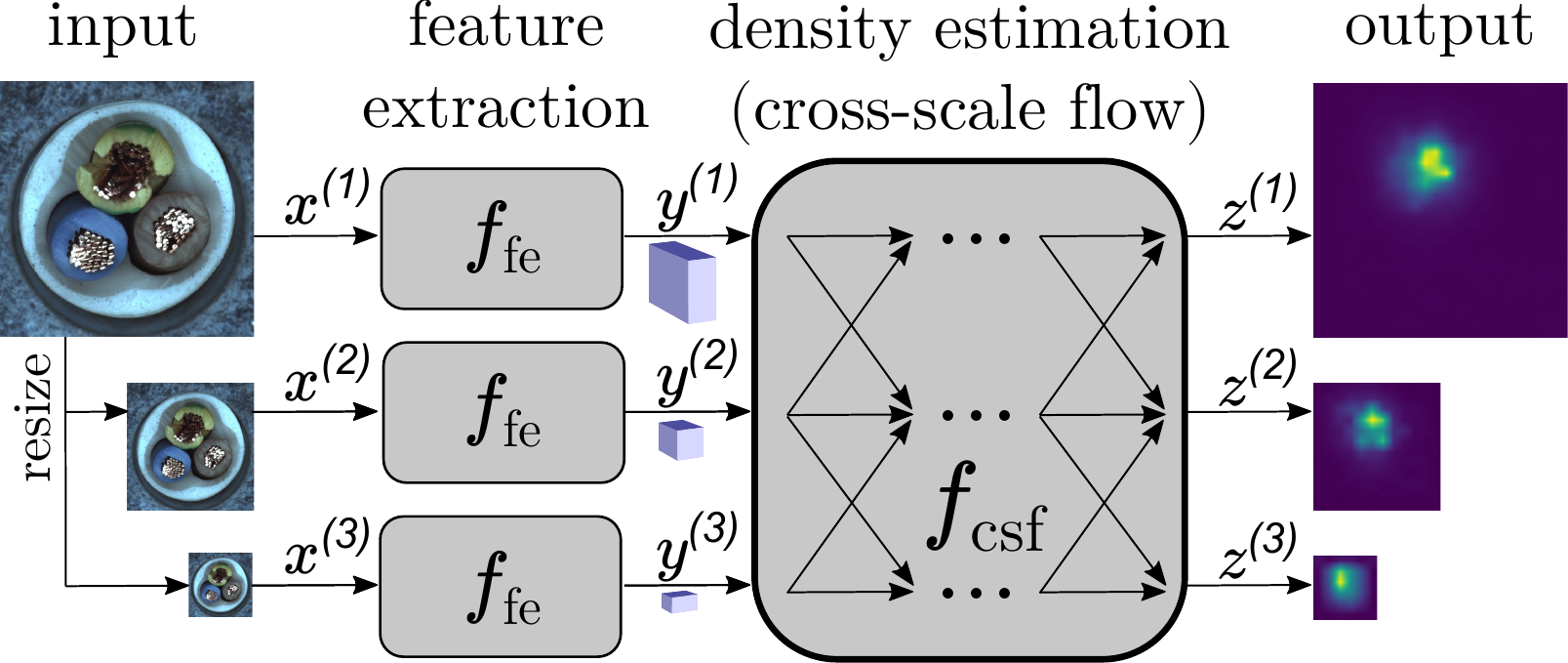} 
 \caption{Our method detects and localizes defects based on the density estimation of feature maps from the differently sized input images. We process the multi-scale feature maps jointly, using a fully convolutional normalizing flow with cross-connections between scales.}
\label{fig:teaser}

    \vspace{-0.5em}
\end{figure}

These terms describe the objective of deciding whether a data sample belongs to the class of the given set $X$ of \textit{normal} (in our case \textit{non-defective}) data.
The problem is interpreted in terms of whether a data sample lies out of the distribution $p_X$ of the set of normal images $X$, also named \textit{out-of-distribution (OOD) detection}.
It is assumed that defects $\bar{X}$ are out-of-distribution, \ie have a small likelihood given $p_X$.
We propose a method that models the distribution on feature level with a normalizing flow.

Most research~\cite{geotrans, csi, deepsvdd} in the AD field focuses on image datasets with high intra-class and high inter-class-variance.
The setting in defect detection is different:
Since the non-defective components are similar to themselves and to the defects, there is a small intra-class and a small inter-class-variance.
Hence, most AD approaches are not suitable for defect detection.
Common approaches based on autoencoders \cite{adae, itae, ae_ssim, memae} or generative adversarial networks (GANs) \cite{anogan, ganomaly, ADGAN} perform poorly in this setting, which is described in detail in Section~\ref{related}.
Thus, recent works rely on density estimation of image features obtained from models pretrained on ImageNet~\cite{imagenet}, \eg ResNet~\cite{resnet} or EfficientNet~\cite{efficientnet}.
However, either information is lost due to the averaging of feature maps~\cite{differnet} or strong statistical priors are required limiting their flexibility in density estimation \cite{rippel, padim}.
To alleviate these issues, we propose a normalizing flow (NF) that is able to process multi-scale feature maps to estimate their density, as shown in Figure~\ref{fig:teaser}.
NFs are generative models that transform the training set distribution $p_X$ to a latent space with a predefined distribution $p_Z$ via maximum-likelihood-optimization.
In contrast to other generative models, for instance VAEs \cite{vae} and GANs \cite{gan, toadgan}, the likelihoods of latent space vectors in NFs are directly interpreted as likelihoods of the input data, since the network maps bijectively.
Thus, the regions in the latent space with high likelihood represent the normal examples while defective examples are projected to latent variables outside of the learned distribution.
Conversely, the injective mapping of autoencoders potentially results in projecting untrained anomalies to indeterminate latent space regions, which may overlap with the regions of the normal samples.

However, applying NFs to images for OOD detection is not straightforward as shown by Kirichenko et al.~\cite{nf_ano_critic}.
With RGB data, the network fails to learn a useful distribution, focusing on local pixel correlations instead of semantics.
For this reason, we perform the density estimation on feature maps obtained by pretrained feature extractors which provide compressed semantic information.
Our \emph{cross-scale flow (CS-Flow)} simultaneously processes the features of the image at different scales by propagating them in parallel through the NF while interacting with each other.
Keeping in mind that the discriminability regarding defectiveness is unknown during training, our model utilizes the full potential of the information and correlations in both local and global contexts to learn the distribution precisely to identify defective examples.
In addition to identification, the fully convolutional architecture also preserves spatial arrangement which allows for a visualization of the defective regions on the image.
In contrast to models using densely connected layers and thus many parameters \cite{differnet}, our approach still achieves good performance even with a low number of training samples.

We summarize our contributions as follows:
\begin{packed_enum}
    \item Our novel \emph{cross-scale normalizing flow (CS-Flow)} detects defects by jointly estimating likelihoods on multi-scale feature maps. 
    \item Our method maintains the image structure to obtain an interpretable latent space, which enables precise defect detection.
    \item We set a new state-of-the-art in image-level defect detection on the MVTec AD and Magnetic Tile Defects dataset.
    \item Code is available on GitHub\footnote{\url{https://github.com/marco-rudolph/cs-flow}}.
\end{packed_enum}


\begin{figure}
\centering
  \includegraphics[width=0.49\textwidth]{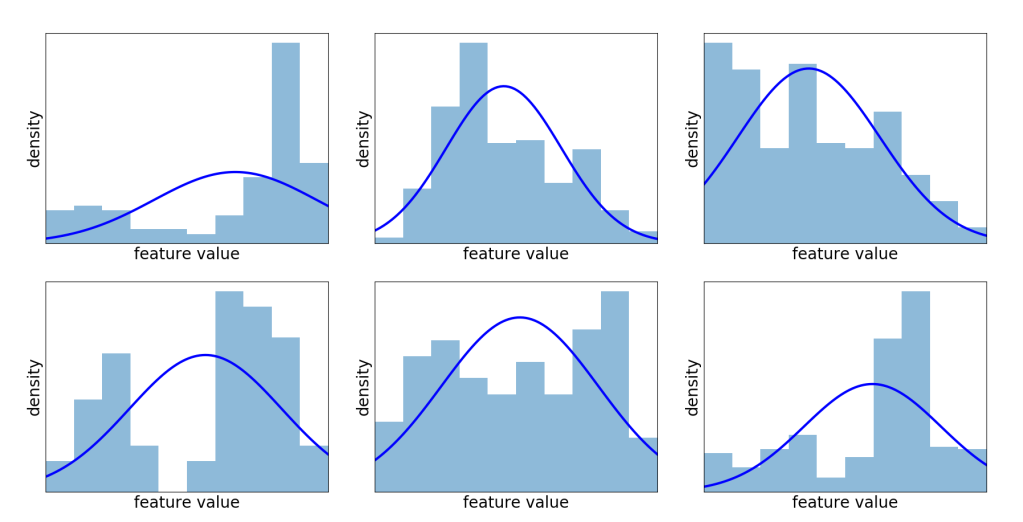} 
 \caption{Histogram of different features from MVTec AD images extracted with EfficientNet~\cite{efficientnet}. Each histogram contains the values from the same position of one feature map.
 The blue line shows the best fitting normal distribution.
 Assuming a normal distribution of the features, as done by \cite{rippel, padim}, appears to be insufficient to capture the feature distribution.
 }
    \vspace{-1.5em}
\label{fig:feat_dist}
\end{figure}

\begin{figure*}
\centering
  \includegraphics[width=0.9\textwidth]{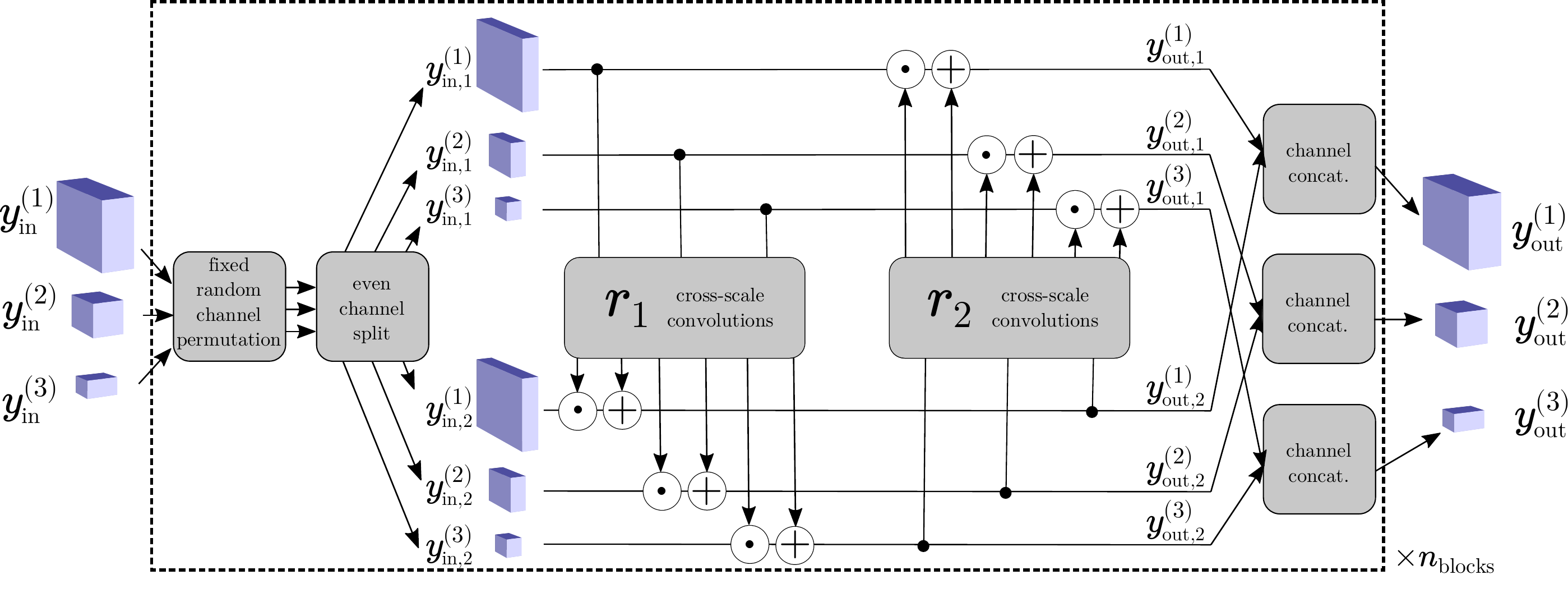} 
 \caption{Architecture of one block inside the normalizing flow: After a fixed random permutation, every input tensor is split into two parts across the channel dimension where each ensemble is used to estimate scale and shift parameters that transform the respective counterpart.
 Symbols $\odot$ and $\oplus$ denote element-wise multiplication and addition, respectively.
 }
\label{fig:block}
    \vspace{-1em}
\end{figure*}

\section{Related Work}
\label{related}
In the following, we review previous work in the field of anomaly detection and normalizing flows as the basis of our methodology.

\subsection{Anomaly Detection}
State-of-the-art work can be roughly divided into approaches that are based on generative models or pretrained networks.
Alternative methods that do not fall into one of these categories are described separately.

\subsubsection{Generative Models}
Many anomaly detection methods are based on generative models, such as autoencoders \cite{ae_lecun,vae,sae} and GANs~\cite{gan}, which are optimized to generate the normal data.
These approaches detect anomalies by the inability of the generative model to reconstruct them.
In the simplest case, the input and the reconstruction of an autoencoder is compared \cite{adae}.
In this context, a high reconstruction error is interpreted as an indicator of an anomaly.
Bergmann \etal \cite{ae_ssim} replace the common $l_2$ error with SSIM to have a better metric for visual similarity.
Gong \etal \cite{memae} use memory modules in the latent space to prevent the autoencoder from generalizing to anomalous data.
Zhai \etal \cite{dsebm} combine energy-based models and regularized autoencoders to model the data distribution.
Denoising autoencoders are used by Huang \etal \cite{itae} by letting autoencoders learn to restore transformed images.

Similar to the decoding part of autoencoders, generators of GANs are utilized for anomaly detection.
Schlegl \etal \cite{anogan} propose to learn an inverse generator after training a GAN, utilizing both together for reconstruction and the error consideration.
A combination of autoencoders and GANs is proposed by Akcay \etal \cite{ganomaly}. 
They apply the autoencoder directly as the GAN's generator to ensure the generation of normal data only.

As shown in Section~\ref{detection}, autoencoders and GANs perform poorly on defect detection tasks.
Since different types of anomalies with individual size, shape and structure have inconsistent characteristics regarding reconstruction errors, they are not widely applicable.
For example, structures with high frequency cannot be represented and reconstructed accurately in general and small defect areas cause smaller errors.

\subsubsection{Methods Based on Pretrained Networks}
Instead of working on the image directly, many methods perform defect detection on features of pretrained networks.
Pretraining on a large-scale database, such as ImageNet, ensures the extraction of universal features that are expected to differ in the presence of defects.
In this way, discriminant features are considered which cannot be learned from non-defective data, since they do not necessarily occur in it.
Detecting defects in the feature space commonly is done using traditional statistical approaches.

Andrews \etal \cite{andrews} fit a one-class Support Vector Machine to the feature distribution.
Rippel \etal \cite{rippel} model the features as an unimodal Gaussian distribution and utilizes the Mahalanobis distance as scoring function.
This approach was further refined by Defard \etal~\cite{padim} by applying it to image patches utilizing feature maps at different semantic levels.
However, these approaches are limited to normal distributions which 
are inappropriate in many cases as shown in Figure~\ref{fig:feat_dist}.
In contrast, we do not assume any predefined feature distribution, but learn the true distribution via maximum likelihood estimation (MLE).
Assuming that distances within the feature space are semantically expressive, the distance to the nearest neighbour is used as an anomaly score in \cite{nazare}. 
The only deep-learning-based image feature density estimation method by Rudolph \etal \cite{differnet}, which is the most comparable to our work, is also based on normalizing flows.
However, they do not process full-sized feature maps, but rather vectors after applying average pooling.
As a result, important contextual and positional information is lost.
The authors partially compensate this weakness by passing 64 different rotations of each image through the network, which, however, significantly increases computational complexity.
In contrast, our method utilizes the fine-grained information of the full-sized feature maps while requiring only a single pass and outperforms DifferNet~\cite{differnet} in almost all experiments by a large margin.

\subsubsection{Other Approaches}
Besides generative and pretrained models, there are alternative approaches to perform anomaly detection.
Liznerski \etal \cite{xdocc} propose a learnable hypersphere classifier using exemplar outlier exposures as anomaly substitute.
Contrastive learning on augmentations of the same image is used by Tack \etal \cite{csi} by defining in-distribution and out-of-distribution transformations.
In contrast, Golan and El-Yaniv \cite{geotrans} augment images to classify the specific transformation, assuming that this does not work as clearly on anomalies as it does on normal data.

\begin{figure*}
\centering
  \includegraphics[width=0.85\textwidth]{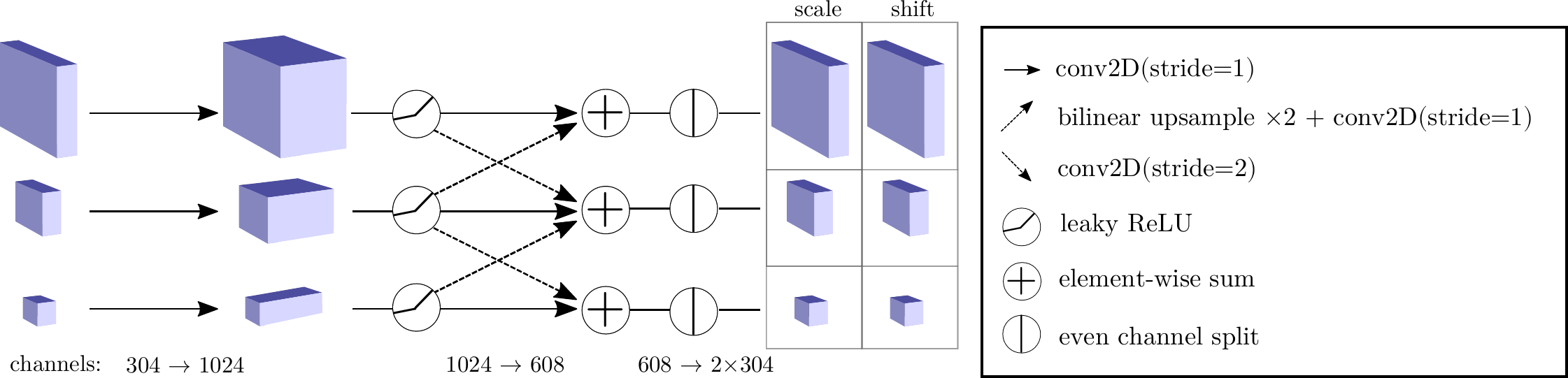} 
 \caption{Architecture of the internal networks $r$ inside the coupling blocks. Convolutions are performed at two levels, with cross-connections between scales at the second level. Feature map resizing is implemented by upsampling and strided convolutions. Aggregation is implemented by summation. The output is split across the channel dimension to obtain the scale and shift parameters.}
    \vspace{-1em}
\label{fig:cross_conv}
\end{figure*}

\subsection{Normalizing Flows}
A normalizing flow (NF) \cite{nf} is a generative model that transforms data into tractable distributions.
Unlike conventional neural networks, their mapping is bijective, which allows them to train and evaluate in both directions~\cite{tomINN}.
The forward pass projects data into a latent space to calculate exact likelihoods for the data given the predefined latent distribution.
Conversely, data sampled from the predefined distribution can be mapped back into the original space to generate data.
Bijectivity and bidirectional execution are ensured by using invertible affine transformations.
There are different types of normalizing flows, which differ in the architecture of the affine transformations in order to efficiently enable the forward or backward direction.
The affine blocks are realized either by learning fixed or autoregressive transformations.
A popular type of autoregressive flows is \textit{MADE} (Germain \etal \cite{germain}).
The density calculation based on the Bayesian chain rule is efficient in this case.
However, sampling is costly.
In contrast, inverse autoregressive flows (Kingma \etal \cite{kingma}) are usually efficient at sampling, but not at computing likelihoods.
Real-NVP \cite{realnvp}, a variant of inverse autoregressive flows, simplifies both passes to be efficient in both directions.
We enhanced Real-NVP to operate on multiple scales that can interact with each other.
This leverages NFs for defect detection by introducing fully convolutional cross-scale flows, whose architecture is explained in detail in Section~\ref{meth_normflow}.

Normalizing flows are successfully used for anomaly detection on non-image data \cite{nf_deep, nf_time_series, nf_trajectory}.
With image data, the problem arises that the network mainly focuses on local pixel correlation without taking semantics into account.
Recent works \cite{differnet, nf_ano_critic} found that semantic information is better captured when working on image features instead of full images.
In contrast to \cite{nf_ano_critic}, we use features from multiple scales and refrain from the usage of fully connected layers and squeeze layers\footnote{Squeeze layers reshape the tensor, \eg by aggregating the channels of 4 neighboring pixels to one pixel with fourfold channel number.}.
In this way, our latent space preserves the spatial arrangement and therefore enables precise defect localization.
Furthermore, we lower the number of parameters which enables us to process high dimensional feature maps and train with few data samples.

\section{Method}
To detect defects in images, we first learn a statistical model of features $\bm{y} \in Y$ of defect-free images $\bm{x} \in X$ similar to DifferNet\cite{differnet}.
During inference, we assign a likelihood to the input image $\bm{x}$
by using a density estimation on image features $\bm{y}$, assuming a low likelihood is an indicator for a defect.
The density estimation is learned via a bijective mapping of the unknown distribution $p_Y$ of the feature space $Y$ to a latent space $Z$ with a Gaussian distribution $p_Z$.
Thus, as shown in Figure~\ref{fig:teaser}, our method is divided into the steps feature extraction $X \rightarrow Y$ and density estimation $Y \rightarrow Z$.

From the input image $\bm{x}$ we extract the features $\bm{y}$ by using a pretrained neural network $f_{\mathrm{fe}}(\bm{x}) = \bm{y}$ which will remain unchanged during training.
To have a more descriptive representation of $\bm{x}$, feature maps of different scales are included in $\bm{y}$ via extracting features from $s$ different resolutions of the image.
In contrast to \cite{differnet}, our proposed NF-architecture is able to perform density estimation on different scaled full-sized feature maps in parallel instead of on concatenated feature vectors.
Thus, important fine-grained positional and contextual information is maintained.
We define $\bm{y} = [y^{(1)}, ..., y^{(s)}]$ with $y^{(i)}$ as the 3D feature tensor of the image $x^{(i)}$ at scale $i \in \{1, ..., s\}$.
Our proposed cross-scale-flow $f_{\mathrm{csf}}$ transforms the feature tensors bijectively and in parallel to 

\begin{equation}
f_{\mathrm{csf}}(y^{(1)}, ..., y^{(s)}) = [z^{(1)}, ..., z^{(s)}] = \bm{z} \in Z
\end{equation}
with the same dimensionality\footnote{For better readability, in the following $\bm{z}$ without any index represents a vector which is the concatenation of the flattened tensors $[z^{(1)}, ..., z^{(s)}]$.} as $\bm{y}$. The likelihood $p_Z(z)$ is measured according to the target distribution which in our case is a multivariate standard normal distribution $\mathcal{N}(0,\,I)$.

We use the likelihood of $p_Z(\bm{z})$ to decide whether $\bm{x}$ is anomalous according to a threshold $\theta$:

\begin{equation}
    \mathcal{A}(\bm{x}) = 
      \begin{cases}
        1 & \text{for } p_Z(\bm{z}) < \theta \\
        0 & else
      \end{cases}
      .
\end{equation}

\subsection{Cross-Scale Flow}
\label{meth_normflow}
We extend the traditional normalizing flows with our novel cross-scale flow to allow for effective defect detection on images.
It processes feature maps of different sizes which interact with each other.
In this way, information between the scales is shared to obtain a likelihood for the compound of $\bm{y} = [y^{(1)}, ..., y^{(s)}]$.
Moreover, we design it fully convolutional and preserve the spatial dimensions.
This allows to determine the positions of the anomalies in $Z$ as shown in Section~\ref{loc}.
An additional benefit of our approach compared to \cite{differnet} is a practicable handling of very high-dimensional input spaces while having few training samples as shown in Section~\ref{experiments}.

The cross-scale flow is a chain of so-called \textit{coupling blocks}, each performing affine transformations. 
As basis for the frame architecture of the coupling block we chose Real-NVP \cite{realnvp}.
The detailed structure of one block with $s=3$ is shown in Figure~\ref{fig:block}. 
Inside, each input tensor $y^{(i)}_{\mathrm{in}}$ is first randomly permuted and evenly split across its channel dimension into the two parts $y^{(i)}_{\mathrm{in}, 1}$ and $y^{(i)}_{\mathrm{in}, 2}$. 
These parts manipulate each other by regressing element-wise scale and shift parameters which are successively applied to their respective counterparts to obtain the output $[y^{(i)}_{\mathrm{out}, 1}, y^{(i)}_{\mathrm{out}, 2}]$.
The scale and shift parameters are estimated by coupling block-individual subnetworks $r_1$ and $r_2$ whose output is split into $[s_1, t_1]$ and $[s_2, t_2]$ and is then used as follows:
\begin{equation}
 \begin{aligned}
\bm{y}_{\mathrm{out}, 2} = \bm{y}_{\mathrm{in}, 2} \odot e^{\gamma_1 \, s_1(\bm{y}_{\mathrm{in}, 1})} + \gamma_1 \, t_1(\bm{y}_{\mathrm{in}, 1})  \\
\bm{y}_{\mathrm{out}, 1} = \bm{y}_{\mathrm{in}, 1} \odot e^{\gamma_1 \,  s_2(\bm{y}_{\mathrm{\mathrm{out}}, 2})} + \gamma_2 \, t_2(\bm{y}_{\mathrm{\mathrm{out}}, 2}),
\end{aligned}
\label{aff}
\end{equation}
with $\odot$ as the element-wise product. 
To initialize the model in a stable way, we introduce the learnable block-individual scalar coefficients $\gamma_1$ and $\gamma_2$.
They are initialized to $0$ and thus cause $y_{\mathrm{out}}=y_{\mathrm{in}}$.
The affinity property is preserved by having non-zero scaling coefficients with the exponentiation  in Equation~\ref{aff}.
The internal networks $r_1$ and $r_2$ do not need to be invertible and can be any differentiable function, which in our case is implemented as a fully convolutional network that regresses both components by splitting the output (see Figure~\ref{fig:cross_conv} for details of the architecture).
Features are processed with one hidden layer per scale on which the number of channels is increased.
Motivated by HRNet \cite{hrnet}, we adjust the size of individual feature maps of different scales by bilinear upsampling or strided convolutions before aggregation by summation.

We apply soft-clamping to the scale components $s$, as proposed by Ardizzone \etal \cite{cinn}, to preserve model stability in spite of the exponentiation.
This clamping is applied as the last layer to the outputs $s_1$ and $s_2$ by the activation
\begin{equation}
\sigma_{\alpha}(h) = \frac{2\alpha}{\pi}\arctan{\frac{h}{\alpha}}.
\end{equation}
This prevents extreme scaling components by restricting the values to the interval $(-\alpha, \alpha)$.

\subsection{Learning Objective}
During training, we want the cross-scale flow $f_{\mathrm{csf}}$ to maximize the likelihoods of feature tensors $p_Y(\bm{y})$ which we obtain by mapping them to the latent space $Z$ where we model a well-defined density $p_Z$.
Using the change-of-variables formula Eq.~\ref{eqn:change_of_variables} and $\bm{z}=f_{\mathrm{NF}}(\bm{y})$, this likelihood is defined by
\begin{equation}
\label{eqn:change_of_variables}
    p_Y(\bm{y}) = p_Z(\bm{z}) \abs{
    \det{
    \frac{\partial \bm{z}}
        {\partial \bm{y}}
    }}
    .
\end{equation}
We optimize the log-likelihood, since it is equivalent and more convenient for a density $p_Z$ of a Gaussian distribution.
Thus, we formulate our objective as the minimization of the negative log-likelihood $-\log{p_Y(\bm{z})}$:
\begin{equation}
 \begin{aligned}
    \log{p_Y(\bm{y})} = \log{p_Z(\bm{z})}  + \log{\abs{
    \det{
    \frac{\partial \bm{z}}
        {\partial \bm{y}}
    }}}\\
    \mathcal{L}(\bm{y}) = -\log{p_Y(\bm{y})} =  \frac{\norm{\bm{z}}_2^2}{2}    - \log{\abs{
    \det{
    \frac{\partial \bm{z}}
        {\partial \bm{y}}
    }}}
    .
 \end{aligned}
 \label{formula:loglikelihood}
\end{equation}
 with $\abs{\det{\frac{\partial \bm{z}} {\partial \bm{y}}}}$ denoting the absolute determinant of the Jacobian.
The logarithm of this term simplifies in our case to the sum of all values of $s$ since the Jacobian of the element-wise product operator in Equation~\ref{aff} is a diagonal matrix.
The training is conducted over a fixed number of epochs.
To stabilize it further, we limit the $l_2$-norm of the gradients to 1.
Section~\ref{implementation} describes the training in more detail.

\subsection{Localization}
\label{loc}
In previous work \cite{differnet}, the latent space of the normalizing flow has only been used such that all entries of $z$ are considered to produce a score at the image level.
Since our method processes feature maps fully-convolutional, positional information is preserved.
This allows for the interpretation of the output in terms of the likelihood of individual image regions, which in our application is the localization of the defect.

Analogous to the definition of the anomaly score of the entire image, we define an anomaly score for each local position $(i, j)$ of the feature map $y^s$ by aggregating the values along the channel dimension with $\norm{z^{s}_{i, j}}^2_2$.
Thus, we can localize the defect by marking image regions with high norm in the output feature tensors $z^{s}$.

\section{Experiments}
\label{experiments}

\begin{table*}
\begin{center}
\hskip-0.5cm
\footnotesize
\begin{tabular}{c|l|c|c|c|c|c|c|c|c|c|c|}
\cline{2-11}
& Category & ARNet & Geom. & GAN & DSEBM & Mahal. & 1-NN & DifferNet & PaDiM & \textbf{CS-Flow (ours) }\\ 
 & &\cite{itae}& \cite{geotrans} &\cite{ganomaly}  & \cite{dsebm} &  \cite{rippel} & \cite{nazare}& \cite{differnet} & \cite{padim} & (16 shots/full set)\\
\cline{2-11}
& Grid       & 88.3 & 61.9 & 70.8 & 71.7  & 93.7 & 81.8 & 84.0 & - & 93.3 \quad \textbf{99.0}\\
& Leather    & 86.2 & 84.1 & 84.2 & 41.6 &\textbf{100} & \textbf{100} & 97.1 & - & \textbf{100} \quad \textbf{100}\\
& Tile       & 73.5 & 41.7 & 79.4 & 69.0 & \textbf{100} & \textbf{100} & 99.4 & - & 99.9 \quad \textbf{100}\\
& Carpet     & 70.6 & 43.7 & 69.9 & 41.3 & 99.6 & 98.5 & 92.9 &-& \textbf{100} \quad \textbf{100}\\
\rotatebox[origin=c]{90}{\parbox[c]{0cm}{Textures}}& Wood       & 92.3 & 61.1 & 83.4 & 95.2 & 99.3 & 95.8 & 99.8 & - & 99.5 \quad \textbf{100}\\
\cline{2-11}
& Avg. Text. & 82.2 & 59.6 & 77.5 & 63.8 & 98.5 & 96.1 & 94.6 & 99.0 & 98.5 \quad \textbf{99.8}\\
\cline{2-11}
& Bottle     & 94.1 & 74.4 & 89.2 & 81.8 & 99.0 & 99.6 & 99.0  & - & \textbf{100} \quad 99.8\\
& Capsule    & 68.1 & 67.0 & 73.2 & 59.4 & 96.3 & 89.4 & 86.9 & - & 83.1 \quad \textbf{97.1}\\
& Pill       & 78.6 & 63.0 & 74.3 & 80.6 & 91.4 & 79.9 & 88.8 & - & 90.9 \quad  \textbf{98.6} \\
& Transistor & 84.3 & 86.9 & 79.2 & 74.1 & 98.2 & 95.4 & 91.1 & - & 98.0 \quad \textbf{99.3}\\
& Zipper     & 87.6 & 82.0 & 74.5 & 58.4 & 98.8 & 97.1 & 95.1 & - & 95.3 \quad \textbf{99.7} \\
& Cable      & 83.2 & 78.3 & 75.7 & 68.5 & \textbf{99.1} & 95.1 & 95.9 & - & 94.4 \quad \textbf{99.1} \\
\rotatebox[origin=c]{90}{\parbox[c]{0cm}{Objects}}& Hazelnut   & 85.5 & 35.9 & 78.5 & 76.2 & \textbf{100} & 98.2 & 99.3 & - & 97.9 \quad 99.6\\
& Metal Nut  & 66.7 & 81.3 & 70.0 & 67.9 & 97.4 &91.1 & 96.1 & - & \textbf{99.1} \quad \textbf{99.1}\\
& Screw      & \textbf{100} & 50.0 & 74.6 & 99.9 & 94.5 & 91.4 & 96.3 & - & 65.2 \quad 97.6\\
& Toothbrush & \textbf{100} & 97.2 & 65.3 & 78.1 & 94.1 & 94.7 & \textbf{98.6} & - & 85.6 \quad 91.9\\
\cline{2-11}
& Avg. Obj.  & 84.8 & 71.6 & 75.5 & 74.5 & 96.9 & 93.2 & 94.7 & 97.2& 91.0 \quad \textbf{98.2} \\
\cline{2-11}
& \textbf{Average} & 83.9 & 67.2 & 76.2 & 70.9 & 97.5 & 93.9 & 94.7 & 97.9 & 93.5 \quad \textbf{98.7}\\
\cline{2-11}
\end{tabular}
\end{center}
\caption{Area under ROC in \% for detecting defects of all categories of MVTec AD \cite{mvtec} on image-level grouped into textures and objects. Best results are in bold.
\textit{16 shots} denotes that a subset of only $16$ random images per category was used in training. Beside the average value, detailed results of PaDiM~\cite{padim} were not provided by the authors. 
}
\label{table:mvtec}
\end{table*}

\begin{table}
\small
\begin{center}
\footnotesize
\begin{tabular}{|l|c|}
\hline
Method & AUROC $[\%] \uparrow$ \\
\hline
Geom. \cite{geotrans} & 75.5 \\
GANomaly \cite{ganomaly} & 76.6 \\
DSEBM \cite{dsebm} & 57.2 \\
Mahalanobis \cite{rippel} & 98.0 \\
1-NN \cite{nazare} & 97.8 \\
DifferNet \cite{differnet} & 97.7 \\
PaDiM \cite{padim} & 98.7 \\
\textbf{CS-Flow (ours)} & \textbf{99.3} \\
\hline
\end{tabular}
\end{center}
\caption{Area under ROC in \% for detecting anomalies on MTD.}
    \vspace{-0.5em}
\label{table:magnets}
\end{table}

\subsection{Datasets}
We evaluate our method on a wide range of realistic defect detection scenarios to demonstrate the advantage of our contributions and the superiority over previous approaches.
For this purpose, we measure the performance on the challenging and diverse MVTec AD \cite{mvtec} and Magnetic Tile Defects (MTD) \cite{magnets} datasets.

MVTec AD comprises 10 object and 5 texture classes with overall 3629 defect-free training and 1725 testing images.
Each class contains 60 to 320 high-resolution images with a range from $700 \times 700$ to $1024 \times 1024$ pixels.
The test set includes defects of different sizes, shapes and types such as cracks, scratches and displacements, with up to 8 different defect types per class and 70 defect types in total.
To the best of our knowledge, MVTec AD acts currently as the only dataset with multi-object and multi-defect-data for anomaly detection.

As a common choice, we also evaluate on the MTD dataset, which includes gray-scale images of magnetic tiles with and without defects.
The contained defects, \eg breaks and blowholes, can cause problems in engines due to an unequal magnetic potential.
It is notable that this dataset shows a large variance within the defect-free examples due to the differences in illumination and other non-defect characteristics.
Following \cite{differnet}, we use all 392 defect images and one fifth of the 952 defect-free images for testing and train on the remaining defect-free data.

\subsection{Implementation Details}
\label{implementation}
We utilize the output of layer 36 of EfficientNet-B5~\cite{efficientnet} as the feature extractor for all experiments as it provides feature maps having a good balance between level of feature semantic and spatial resolution.
The feature extractor remains fixed during training after being pretrained on ImageNet~\cite{imagenet}.
For MVTec AD, we use features at $s=3$ scales with input image sizes of $768\times768$, $384\times384$ and $192\times192$ pixels - resulting in feature maps with spatial dimensions $24\times24$, $12\times12$ and $6\times6$ and each $304$ channels.
Due to the smaller original image size of MTD samples, we resized the images to $384\times384$, $192\times192$ and $96\times96$ pixels.
We use $n_{\mathrm{blocks}} = 4$ coupling blocks inside CS-Flow using $3 \times 3$ convolutional kernels in internal networks for the first 3 blocks and $5 \times 5$ kernels for the last block.
The clamping parameter is set to $\alpha=3$ and the negative slope of the leaky ReLU is set to $0.1$.
For optimization, we use Adam~\cite{adam} with a learning rate of $2 \cdot 10^{-4}$, a weight decay of $10^{-5}$ and momentum values $\beta_1=0.5$ and $\beta_2 = 0.9$.
We train our models with a batch size of 16 for a fixed number of 240 epochs for MVTec AD and 60 epochs for MTD, respectively, since there is no validation set to define a stopping criterion.
A training run of one class of MVTec AD takes about 45 minutes on average using a NVIDIA  RTX  2080  Ti.

\begin{figure}
\centering
  \includegraphics[width=0.46\textwidth]{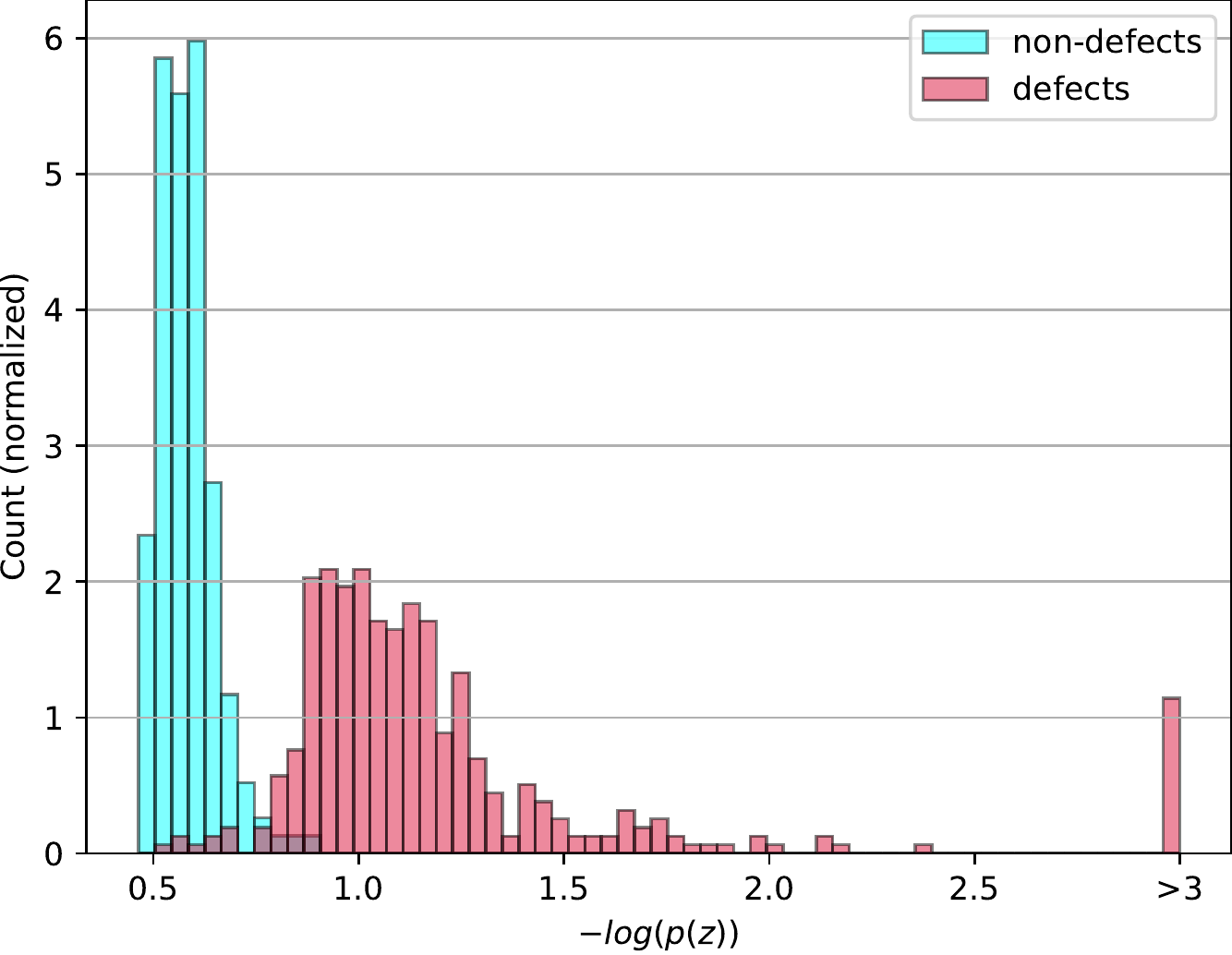} 
 \caption{Distribution of negative log-likelihood for test images of MTD as a normalized histogram. By this criterion, the defective samples are almost completely separable from the non-defective samples. Note that for clarity, the rightmost bar summarizes all scores above 3.}
    \vspace{-1.5em}
\label{fig:scores}
\end{figure}

\begin{figure}
\centering
  \includegraphics[width=0.4\textwidth]{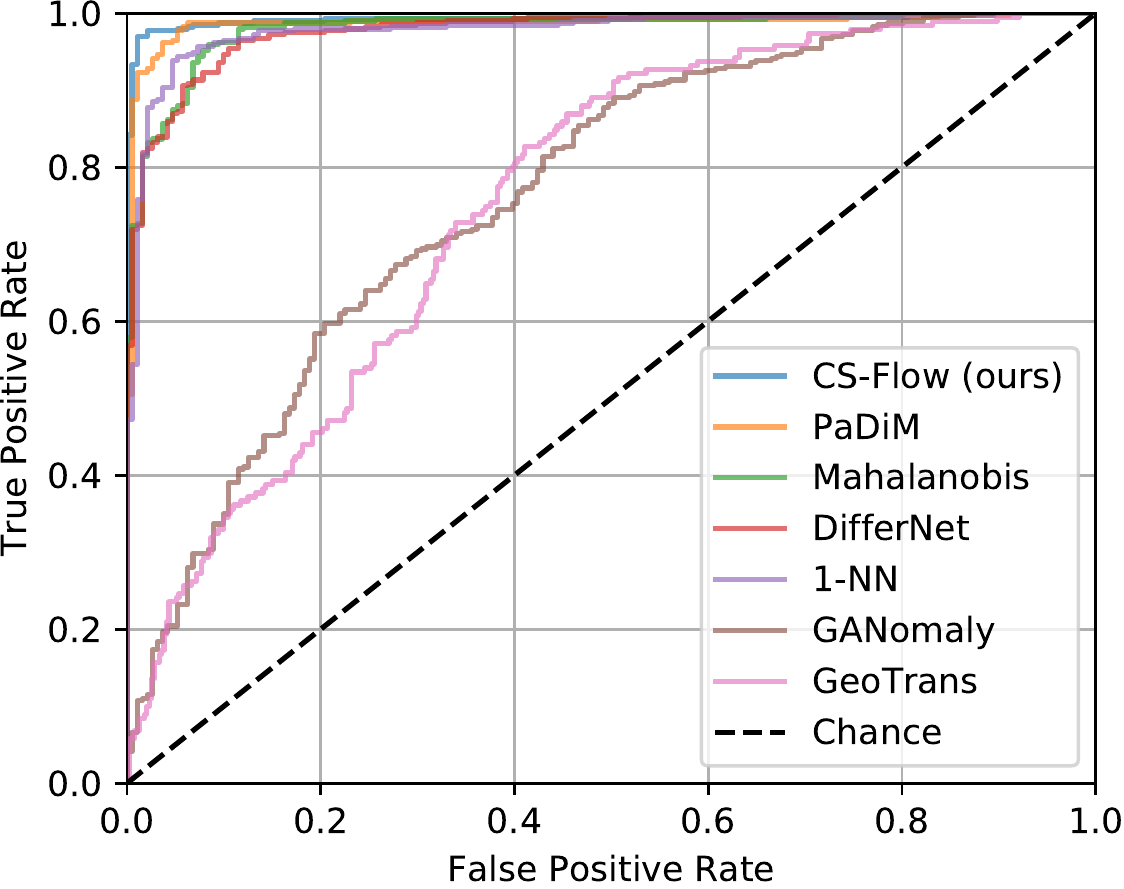} 
 \caption{Comparison of defect detection performance of different methods on MTD. The graphs are the ROC-Curves of the individual methods. Best viewed in color.}
    \vspace{-1.5em}
\label{fig:magnetroc}
\end{figure}

\subsection{Detection}
\label{detection}
In order to measure and compare the defect detection performance of our models, we follow \cite{differnet} and calculate the area under ROC (AUROC) at image-level on the respective test sets.
The ROC (\textit{Receiver Operating Characteristics}) curve relates the true positive rate to the false positive rate with respect to a parameter (in our case the threshold $\theta$).
Thus, it is invariant to the ratio of anomalies in the set and is therefore representative for realistic settings.
Table~\ref{table:mvtec} shows the defect detection performance of our method and other state-of-the-art works on the individual categories of MVTec AD.
For a fair comparison, we evaluated \cite{nazare} and \cite{rippel} on the same multi-scale features as our method which improved their performance in every case.
Here, we averaged the feature maps of different scales individually, resulting in a feature vector with $3 \cdot 304 = 912$ dimensions.
Note that PaDiM~\cite{padim} is originally based on EfficientNet-B5~\cite{efficientnet}.
Since the results of \cite{differnet} dropped heavily with this backbone, we report the paper-given results with AlexNet~\cite{alexnet}.
We outperform or match the competitors on 12 of 15 categories with an average AUROC of $98.7\%$, which considerably closes the gap to the optimum of 100\% compared to competitors.
CS-Flow works reliably on a wide range of defects having an AUROC over 97\% in 14 of 15 categories.
Our method remains competitive when training on only 16 samples per category, with even showing roughly the same performance on the texture categories.

We also set a new state of the art of 99.3\% AUROC on MTD as shown in Table~\ref{table:magnets}.
As shown in Figure~\ref{fig:scores}, the likelihood assigned by our model clearly distinguishes the defective from the non-defective parts, with only a few exceptions.
Being just $0.7\%$ AUROC close to an optimal ROC, we want to emphasize that in this metric a margin of a few percents compared to competitors is a relatively strong increase in performance as visualized in Figure~\ref{fig:magnetroc}.

\subsection{Localization}
\begin{figure*}
\centering
  \includegraphics[width=0.90\textwidth]{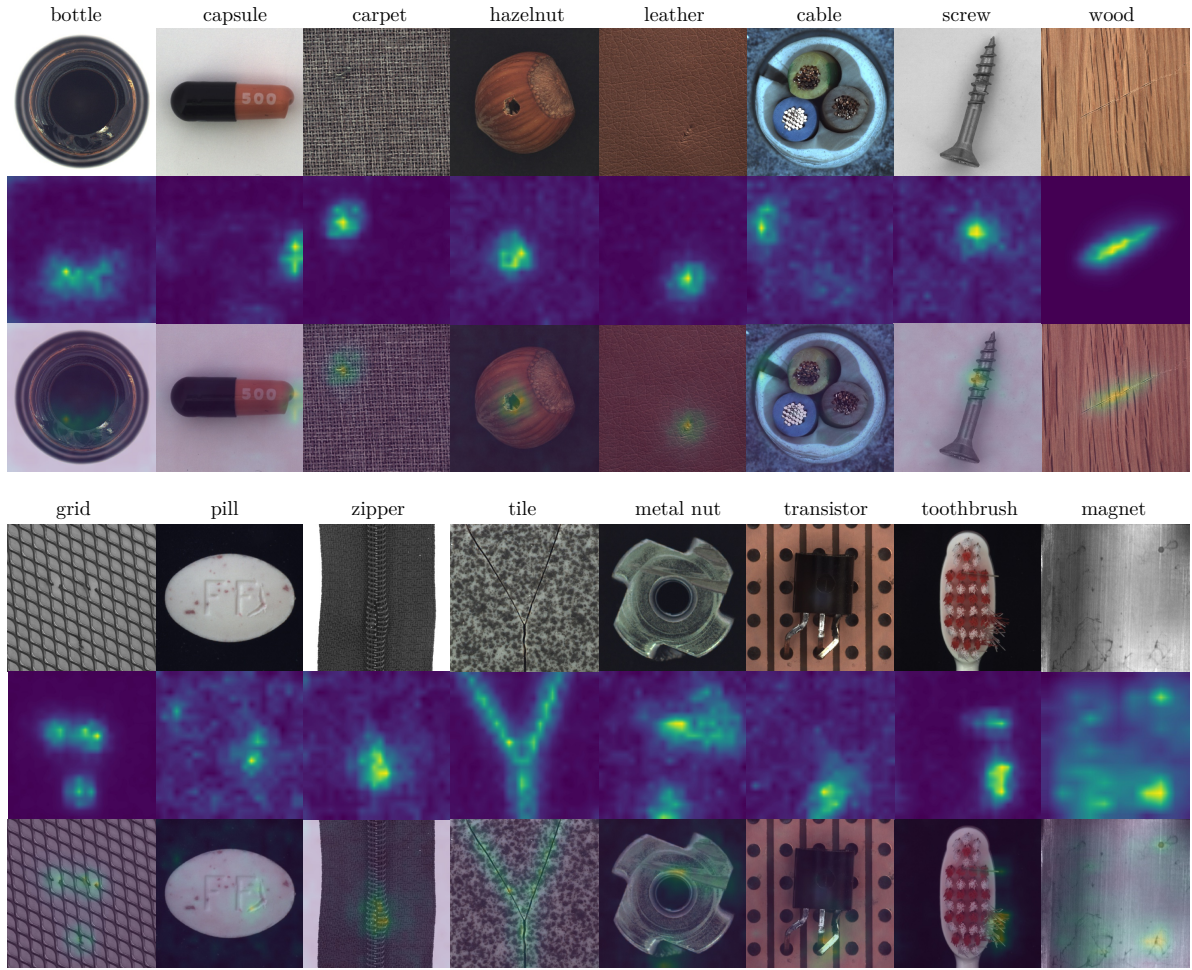} 
 \caption{Defect localization of one defective example per category of MVTec AD and MTD. The rows each show the original image, the localization and the overlay of both images, from top to bottom. The localization maps show the sum of squares along the channel dimension of the networks output at the highest scale.}
    \vspace{-1em}
\label{fig:localization}
\end{figure*}

Although the objective of our approach is to detect defects on image level, it can also be used to localize defective regions in images, due to its global and local feature preserving nature.
In this section we study the localization, as described in Section~\ref{loc}.
Our goal is to give a quick visual feedback to an operator.
Figure~\ref{fig:localization} shows the visualization of the highest scale outputs $z^{(1)}$. 
These were scaled up with a bilinear interpolation after summing up the squared values along the channel dimension.
It can be seen that the magnitude of the output values is directly related to the occurrence of anomalous regions at the respective position.
Therefore, our method localizes anomalies of various sizes with respect to color, pattern and shape.
Except for dilations due to the convolutional receptive field, defective regions are determined properly.
We do not aim to provide pixel-precise segmentations as the method is not optimized for it and processes small-resolution feature maps.
Nevertheless, this visualization helps in the interpretation of the output in practice to quickly find or assess the potential error.
We refer to the supplemental material for more detailed analysis of the localization.

\begin{table}
\small
\begin{center}
\footnotesize
    \begin{tabular}{|l|  c|}
    \hline
    Method & AUROC $[\%]\uparrow$  \\ \hline 
    single scale NF ($768\times768$) & 97.8\\
    single scale NF ($384\times384$) & 96.8\\
    single scale NF ($192\times192$) & 96.1\\
    separate multi-scale & 98.2\\
    concat multiscale & 98.0 \\
    \textbf{CS-Flow (ours)} & \textbf{98.7}\\
    \hline
    \end{tabular}
\end{center}
    \caption{Ablation study on MVTec AD with varying strategies regarding the usage of scales.}
    \vspace{-0.5em}
    \label{table:ablation}
\end{table}

\begin{table}
\small
\begin{center}
\footnotesize
\begin{tabular}{ |c|c|c|c|c|c|c|} 
    \hline
    $n_{blocks}$ & 1 & 2& 3& 4& 5 & 6\\
    \hline
    AUROC $[\%] \uparrow$ & 94.6 & 97.8 & 98.5 & \textbf{98.7} & \textbf{98.7} & 98.6 \\
\hline
\end{tabular}
\end{center}
    \caption{Ablation study on MVTec AD for a different number of coupling blocks.}
    \vspace{-1.5em}
    \label{table:cc}
\end{table}

\label{results}
\subsection{Ablation Studies}
To quantify the influence of the individual design decisions of our model,
we report results obtained when varying the hyperparameters of our method.
Table \ref{table:ablation} shows the results of these experiments.
We measure the impact of the multi-scale approach on the defect detection performance.
To this end, we train models on feature maps from one of the three scales at a time (denoted as \textit{single scale NF}).
The results confirm that the features of a single scale are weaker with respect to the discriminability between defective and non-defective samples.
Furthermore, we set another baseline by adding the log likelihoods provided by the networks from every scale (denoted as \textit{separate multi-scale}).
The increase in AUROC compared to the individual performance for the single scale models demonstrates that the features of different scales complement each other well to obtain a more robust score.
Nevertheless, this method is $0.5\%$ AUROC below the performance of our joint training of the individual scales with CS-Flow.
To test our architecture against a naive approach of joint training, we feed a single-scale NF the concatenation of differently sized feature maps along the channel dimension after upscaling each of them to the highest feature map size with bilinear interpolation, comparable to \cite{padim}. 
This setup (denoted as \textit{concat mutiscale}) results in a performance drop of $0.7\%$, which justifies our cross-convolutional multi-scale procedure.

In another experiment, we studied the influence of the number of coupling blocks. The results in Table \ref{table:cc} show that the performance improves with increasing number of coupling blocks up to $n_{blocks}=4$ and then saturates.

To  test  our  model  on  a  setting with more intra-class variance in the normal data, we additional experiment training simultaneously with all 15 classes  of  MVTec  AD  as  normal  data.
The average detection AUROC is  98.2\% which shows that our model can handle multi-modal distributions.

\section{Conclusion}
We presented a semi-supervised method to effectively detect and localize defects on feature tensors of different scales using normalizing flows.
We utilize the context within and between multi-scale feature maps by integrating cross-convolution blocks inside the normalizing flow to assign likelihoods and detect unlikely samples as defects.
This addresses weaknesses of previous methods that struggle either due to restrictions of overly simplified data representations or limited distribution models and enables our method to set state-of-the-art performance on MVTec AD and MTD.
In the future, the concept could be refined for video anomaly detection \cite{wentong_ano, wentong_ano2}.

\vspace{-1.0em}
\small{\paragraph{Acknowledgements.}
This work was supported by the Federal Ministry of Education and
Research (BMBF), Germany under the project LeibnizKILabor (grant no.
01DD20003), the Center for Digital Innovations (ZDIN) and the Deutsche Forschungsgemeinschaft  (DFG) under  Germany’s  Excellence  Strategy  within  the  Cluster of Excellence PhoenixD (EXC 2122).
\newpage
\clearpage
{\small
\bibliographystyle{ieee_fullname}
\bibliography{egbib}
}

\end{document}